\documentclass[a4paper]{article}

\usepackage{INTERSPEECH2022}

\title{Effectiveness of text to speech pseudo labels for forced alignment and cross lingual pretrained models for low resource speech recognition}
\name{Anirudh Gupta$^1$, Rishabh Gaur$^1$, Ankur Dhuriya$^1$, Harveen Singh Chadha$^1$, Neeraj Chhimwal$^1$, Priyanshi Shah$^1$, Vivek Raghavan$^2$}
\address{
  $^1$Thoughtworks\\
  $^2$Ekstep Foundation}
\email{anirudh.gupta@thoughtworks.com, vivek@ekstep.org}

\begin{document}

\maketitle
\begin{abstract}
In the recent years end to end (E2E) automatic speech recognition (ASR) systems have achieved promising results given sufficient resources. Even for languages where not a lot of labelled data is available, state of the art E2E ASR systems can be developed by pretraining on huge amounts of high resource languages and finetune on low resource languages. For a lot of low resource languages the current approaches are still challenging, since in many cases labelled data is not available in open domain. In this paper we present an approach to create labelled data for Maithili, Bhojpuri and Dogri by utilizing pseudo labels from text to speech for forced alignment. The created data was inspected for quality and then further used to train a transformer based wav2vec $2.0$ ASR model. All data\footnote{https://github.com/Open-Speech-EkStep/ULCA-asr-dataset-corpus} and models\footnote{https://github.com/Open-Speech-EkStep/vakyansh-models} are available in open domain. 
\end{abstract}

\section{Introduction}

E2E ASR systems have gained immense popularity in the recent past and have repeatedly shown to be superior compared to traditional approaches of ASR systems \cite{hinton}. However, huge amount of labelled data is needed for E2E systems to perform well for a particular language. This makes the process of training models for high resource languages whose data is available in public domain easier. But on the other hand for most of the languages very limited amount of training data is available. And for some languages it is not available at all. Due to this reason not a lot of speech and natural language processing (NLP) tools are available for those languages. Moreover manually labelling data is both time consuming and expensive. There has been an interest among researchers for approaches which can bootstrap the process of data collection for low resource languages using high resource languages. 

Transfer learning techniques for speech recognition aim at effectively transferring knowledge from high-resource languages to low-resource languages and have been extensively studied. In \cite{schultz2001language} the authors talk about a language independent and language adaptive approach for acoustic modelling. When using deep neural networks for speech recognition it was shown in \cite{huang2013cross} that hidden layers can be made language independent while the fully connected layer can be language specific. Similar approaches were discussed in \cite{tong2017multilingual, toshniwal2018multilingual} where efforts were made to train a single neural network for acoustic modelling of multiple languages. A popular paradigm for transfer learning in E2E systems is to pretrain a model on labelled speech from one (or more) high-resource languages and then fine-tune all or parts of the model on speech from the lowresource language. In the recent years semi supervised learning approaches have shown that state of the art models can be trained for as less as $5$ minutes of labelled audio given huge amount of data is used for learning speech representations while pretraining \cite{baevski2020wav2vec}. An approach which uses multilingual finetuning is described in \cite{conneau2020unsupervised}. It shows that low resource languages benefit from multilingual pretraining by learning shared speech representations across languages. 

A sensible approach to train ASR models would be to utilize pretrained models and finetune on less amount of labelled data. First we generate labelled data using forced alignment utilizing pseudo labels from a TTS engine and then finetuning is performed. =. The authors are not aware of any other study reporting WER for Maithili, Bhojpuri and Dogri for speech recognition on continuous speech. 

\section{Approach}

Languages spoken in the South Asian region belong to at least four major language families: Indo-European (most of which belong to its sub-branch Indo-Aryan), Dravidian, Austro-Asiatic, and Sino-Tibetan. Almost one third of our mother-tongues in India ($574$ languages) belong to the Indo-Aryan family of languages - spoken by $73.30$\% of Indians\footnote{https://www.education.gov.in}. Maithili, Bhojpuri and Dogri are also part of the Indo Aryan languages. Additionally, all these three languages are predominantly written in \textit{Devanagri} script. Maithili is mostly spoken in Nepal and northern parts of Bihar, a state in India. Bhojpuri is a dialect of Hindi and is spoken in eastern Uttar Pradesh and parts of Bihar, Jharkhand and Nepal. Dogri is chiefly spoken in Jammu and Kashmir and parts of Punjab and Himachal Pradesh. 

In \cite{maithili} it is deduced that phonetic similarity between Hindi and Maithili is $74$\% by looking at the most common words used in both languages. Dogri language sounds more similar to Punjabi than Hindi. This phonetic similarity and same script  as Hindi for the languages under consideration form the basis for using pseudo labels of text to speech using Hindi models. 
Forced alignment is the process of determining, for each fragment of the transcript, the time interval (in the audio file) containing the spoken text of the fragment. Most tools for forced alignment use speech recognition algorithms. For low resource languages this becomes a big challenge due to non-availability of data. \texttt{espeak}\footnote{http://espeak.sourceforge.net/} is a lightweight TTS engine which uses formant synthesis method to synthesize sounds from phonemes and prosody information. Since all three languages use \textit{Devanagri} script Hindi TTS models were used no new rules were written for text to phoneme conversion. Sound synthesis was also not altered and forced alignment was performed by assuming that phonetic similarity between Hindi and Maithili, Bhojpuri and Dogri would produce good quality aligned segments. 

\section{Experiment Details}

We break down our process in two parts. First part is creation of labelled data and second part is training an ASR model on this data. 

\subsection{Dataset Details}

News on AIR\footnote{https://newsonair.gov.in/} is a news service division of \textit{All India Radio}. Multiple news bulletins are made public in various languages on a daily basis. A lot of radio stations broadcast news in their regional languages and verbatim text of these bulletins is also made public in PDF format. 

\subsubsection{Text Extraction}
The first step in the process is to extract text from PDF bulletins which could then be further used for forced alignment. The text was machine readable but was not in standard encodings. For all the three languages text was in \textit{KrutiDev} encoding. Special care was taken to convert all this text to UTF-8 format. 

\subsubsection{Forced Alignment}
There were some attributes of extracted text that could affect the quality of forced alignment adversely. Some of them being: 
\begin{itemize}
	\item The PDF bulletin contained some other information in English sometimes like time of the day, details of speak name etc. which were not part of the audio bulletin.
	\item Before reading out the news, the speaker used to introduce himself or herself. Text of such portions was also not there in the PDF bulletin. 
	\item Sometimes there were long music interludes in between. 
	\item In some bulletins the text was not machine readable. 
\end{itemize}
First we generate audio segments from text using \texttt{espeak} TTS engine. 
All the three languages Maithili, Bhojpuri and Dogri use \textit{Devanagri} script which is same as Hindi. Also, Maithili and Bhojpuri are phonetically very similar to Hindi since they are spoken in the same regions of India. Support for these languages was not present in the TTS engine so we chose Hindi to generate audio segments for text for these languages.  Dynamic Time Warping (DTW) algorithm is used to align the Mel-frequency cepstral coefficients (MFCCs) representation of the given audio wave and the audio wave obtained by synthesizing the text fragments. 

We addressed the issues in mismatch between audio and text bulletins by manually analysing the aligned segments. The following steps helped to improve alignment quality. 

\begin{itemize}
	\item Text from any other script apart from \textit{Devanagri} was removed before alignment. 
	\item In the beginning of each audio bulleting the news presenters were introducing themselves. We found that these introductions were clubbed with the first or second aligned segments. As a result we discarded the initial five instances of aligned segments while counting them in labelled data. 
	\item Any segments which were greater than 15 seconds were rejected. Most of these segments contained musical interludes. And it was observed that longer segments had a higher chance of bad alignment. 
\end{itemize}

\subsubsection{Aligned Data}

Total duration of dataset combining all languages is $ 156.56$ hours. Language wise data duration after forced alignment is shown in Table~\ref{tab:data}. 

\begin{table}[th]
  \caption{Data after alignment}
  \label{tab:data}
  \centering
  \begin{tabular}{lll}
    \toprule
    \textbf{Language} &  \textbf{Train} & \textbf{Valid} \\
    \midrule
    Maithili                       & $41.3$ & $3.1$           \\
    Bhojpuri                       & $56.1$  & $4.8$               \\
    Dogri & $51$  & $3.2$       \\
    \bottomrule
  \end{tabular}
  
\end{table}

\begin{figure}[t]
	\centering
	\includegraphics[width=\linewidth]{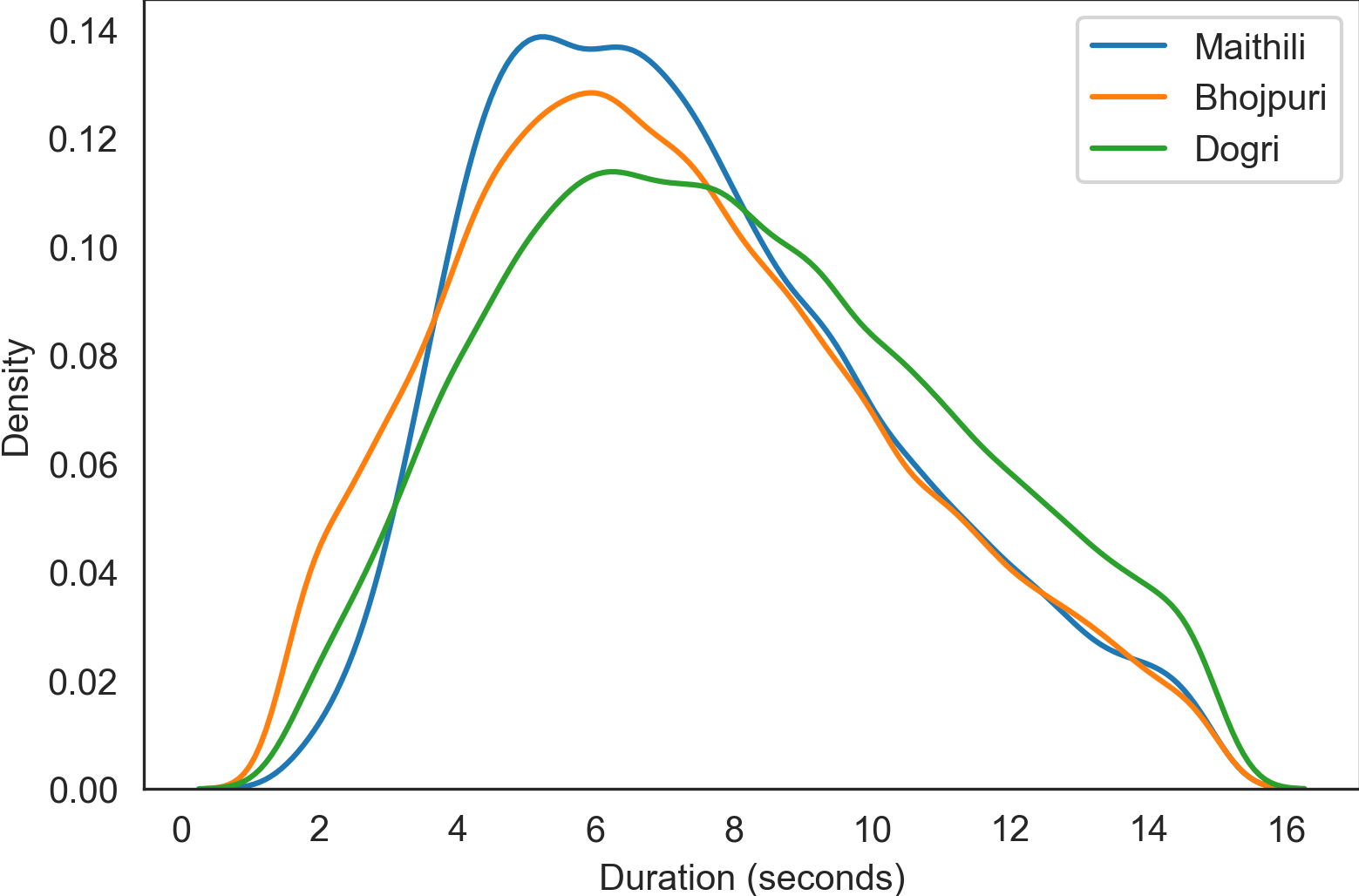}
	\caption{Segment duration distribution}
	\label{fig:speech_production}
\end{figure}

Figure \ref{fig:speech_production} depicts the distribution of audio length segments for the three languages. It can be seen that most of the segment lengths are between $4$ seconds to $10$ seconds. 

To get a better sense of the distribution of aligned data we try to estimate the number of speakers and gender distribution in a given language. Recently a new loss function called the generalized end-to-end (GE2E) loss was introduced which makes the training of speaker verification models more efficient \cite{8462665}. We use an implementation which derives a high level representation of the voice\footnote{https://github.com/resemble-ai/Resemblyzer}. Given an audio, it creates an embedding of $256$ length vector which summarizes the characteristics of spoken voice.

We use the same $256$ length embedding we used for speaker clustering as features for gender identification. Data was created manually and labelled as male or female to train a classifier. Total male audio duration was $17.41$ hours and total female data was $16.25$ hours. To make our model more robust we used clean as well as data with background noise for training. We also included multiple languages which included Hindi,Telugu, Tamil and Kannada. From our analysis a support vector machine \cite{smola} with radial basis function with $\gamma=0.01$ and $C=100$ worked best. We created another test set with new voices and environment in which the labels were balanced. To understand performance on other languages the test set included data from Marathi and Bengali as well. The total duration of the test set was $7.4$ hours. The final accuracy on the test set was $96$\%.

\begin{figure}[t]
	\centering
	\includegraphics[width=\linewidth]{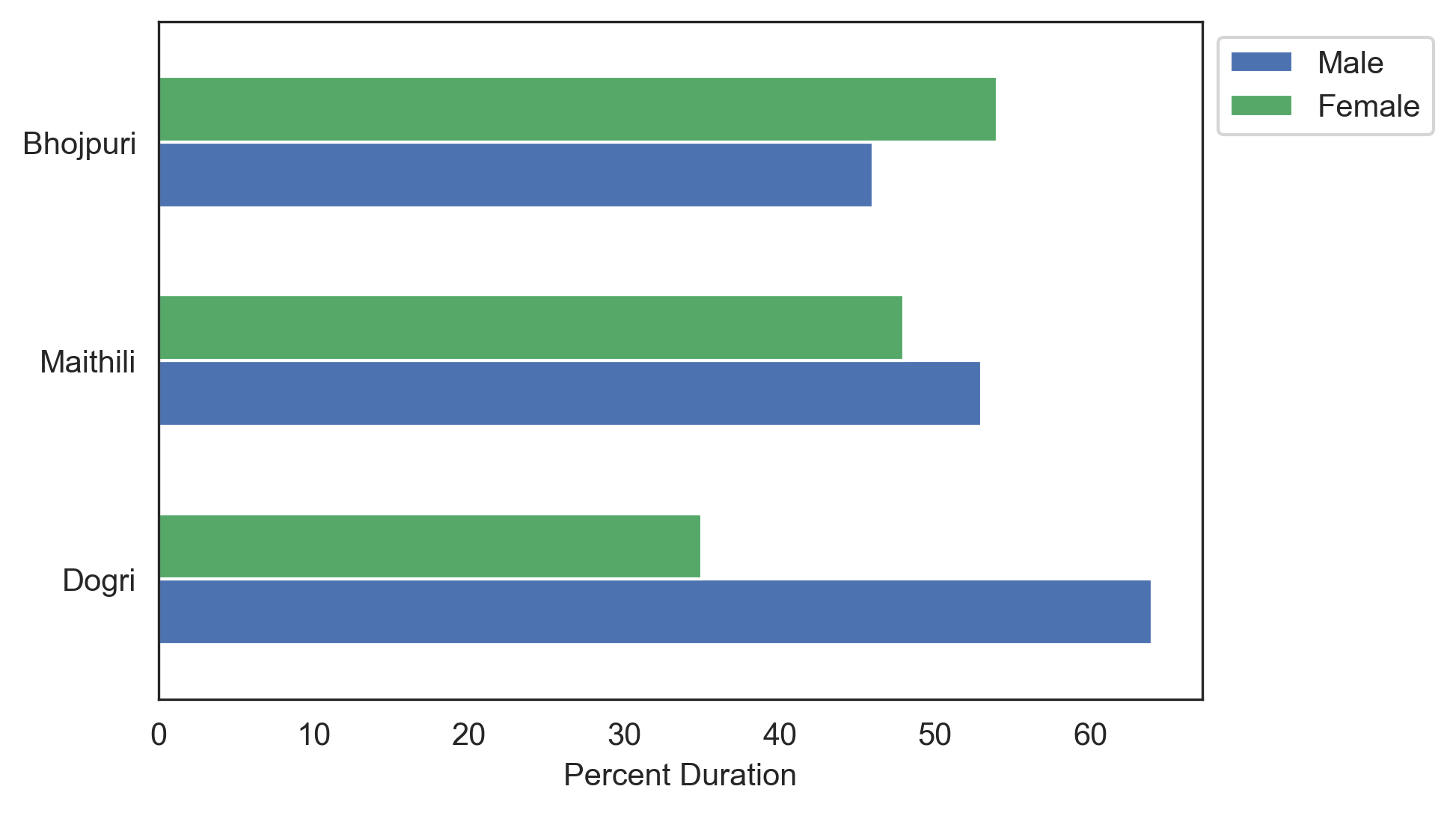}
	\caption{Gender distribution}
	\label{fig:genger}
\end{figure}

\subsection{Model Training}
Wav2vec $2.0$ is a self-supervised pre-training framework for learning speech representations. It is trained by solving a contrastive task over masked latent speech representations and jointly learns quantization of latents shared across languages. \cite{baevski2020wav2vec}. The resulting model is then fine-tuned on labelled data with a CTC loss function \cite{Graves06connectionisttemporal} where the characters of a particular language act as labels. We start with CSRIL-$23$ as the pretrained model. This model is trained on $10,000$ hours of $23$ Indic languages \cite{clsril}. Pretraining data contained $113.8$ hours of Maithili data, $17.1$ hours of Dogri data but no Bhojpuri. This pretrained model is a \text{base} model of the wav2vec $2.0$ architecture and is further finetuned for each language separately. The training is  continued until no improvement in word error rate (WER) on validation set is observed. We use the same recipe as described in \cite{baevski2020wav2vec} for fine-tuning.

\subsection{Results}
The WER  on validation sets are reported in Table \ref{tab:wer}. A $5$-gram KenLM \cite{heafield-2011-kenlm} language model is trained on text from training data and is used while decoding.

\begin{table}[th]
	\caption{Finetuning Results}
	\label{tab:wer}
	\centering
	\begin{tabular}{ll}
		\toprule
		\textbf{Language} &  \textbf{WER}  \\
		\midrule
		Maithili                       & $12.2$  \\
		Bhojpuri                       & $29.7$   \\
		Dogri & $36.5$      \\
		\bottomrule
	\end{tabular}
	
\end{table}

\section{Conclusions}
Getting labelled data for ASR is difficult and even more so for low resource languages. We propose a method to generate labelled data for speech recognition in low resource languages using forced alignment, by leveraging TTS pseudo labels of a phonetically similar languages. It is observed WER is highest for Dogri. This is also due to the fact in that Dogri sounds more like Punjabi than Hindi. We plan to extend this approach for other languages which are phonetically similar to other high resource languages and use the same script. 

\section{Acknowledgements}
All authors gratefully acknowledge Ekstep Foundation for supporting this project financially and providing infrastructure. A special thanks to Dr. Vivek Raghavan for constant support, guidance and fruitful discussions. We also thank Ankit Katiyar, Heera Ballabh, Niresh Kumar R, Sreejith V, Soujyo Sen, Amulya Ahuja and Nikita Tiwari for helping out when needed and infrastructure support for data processing and model training.

\bibliographystyle{IEEEtran}

\bibliography{mybib}

\end{document}